\documentclass[runningheads]{llncs}
\usepackage[T1]{fontenc}
\usepackage{graphicx}
\usepackage{textcomp}
\usepackage{amsmath}
\usepackage{amsfonts}
\usepackage{amssymb}
\usepackage{url}
\usepackage{color}
\usepackage{multirow}
\usepackage{wrapfig}
\usepackage[colorlinks,bookmarksopen,bookmarksnumbered,citecolor=blue, linkcolor=blue, urlcolor=blue]{hyperref}
\usepackage{tikz}
\newcommand*{\circled}[1]{\lower.7ex\hbox{\tikz\draw (0pt, 0pt)%
    circle (.5em) node {\makebox[1em][c]{\small #1}};}}

\usepackage{amsthm}

\begin{document}
\title{One registration is worth two segmentations}
%
%
\author{Shiqi Huang\inst{1,2} \and
Tingfa Xu\inst{2} \and
Ziyi Shen\inst{1} \and Shaheer Ullah Saeed\inst{1} \and Wen Yan\inst{1} \and Dean Barratt\inst{1} \and Yipeng Hu\inst{1}}

%
\institute{University College London, London, UK \and
Beijing Institute of Technology, Beijing, China\\
\email{\{huangsq,ciom\_xtf1\}@bit.edu.cn}\\
\email{yipeng.hu@ucl.ac.uk}}
\maketitle              
\begin{abstract}
The goal of image registration is to establish spatial correspondence between two or more images, traditionally through dense displacement fields (DDFs) or parametric transformations (e.g., rigid, affine, and splines). 
Rethinking the existing paradigms of achieving alignment via spatial transformations, we uncover an alternative but more intuitive correspondence representation: a set of corresponding regions-of-interest (ROI) pairs, which we demonstrate to have sufficient representational capability as other correspondence representation methods.
Further, it is neither necessary nor sufficient for these ROIs to hold specific anatomical or semantic significance. In turn, we formulate image registration as searching for the same set of corresponding ROIs from both moving and fixed images - in other words, two multi-class segmentation tasks on a pair of images. For a general-purpose and practical implementation, we integrate the segment anything model (SAM) into our proposed algorithms, resulting in a SAM-enabled registration (SAMReg) that does not require any training data, gradient-based fine-tuning or engineered prompts. We experimentally show that the proposed SAMReg is capable of segmenting and matching multiple ROI pairs, which establish sufficiently accurate correspondences, in three clinical applications of registering prostate MR, cardiac MR and abdominal CT images. Based on metrics including Dice and target registration errors on anatomical structures, the proposed registration outperforms both intensity-based iterative algorithms and DDF-predicting learning-based networks, even yielding competitive performance with weakly-supervised registration which requires fully-segmented training data.

\end{abstract}
\section{Introduction}
Registering a pair of moving and fixed images is one of the basic tasks in many medical image computing applications~\cite{de2019deep,eppenhof2018deformable,rohe2017svf}, among which the registration-estimated spatial transformation can warp the moving image to be \textit{aligned} with the fixed image. 
Common transformation functions may be parameterized by translation, rotation, scaling, spline coefficients, spatially-sampled control point coordinates, and, most generally, displacement vectors at all voxel locations - a dense displacement field (DDF). Like their \textit{classical} interactive counterparts, recent learning-based registration algorithms have also been formulated to predict these transformation parameters~\cite{detone2016deep}.

Warping images is only one of many possible use cases of the output transformation that represents spatial correspondence. For example, registration-enabled atlas-based label fusion~\cite{yang2018neural}, morphological analysis on predefined anatomical structures~\cite{cerveri2019pair}, surgical navigation~\cite{luebbers2008comparison} and target tracking~\cite{seregni2017hybrid}, require accurate correspondence to be estimated on only one or a small number of regions-of-interest (ROIs).
In fact, few clinical applications require ubiquitous correspondence and, arguably, the registration becomes an \textit{ill-posed} problem when estimating ${N}\times{3}$ displacements at all voxel locations, given ${N}\times{2}$ intensity values from two images, $N$ being the average number of voxels in one image. 



Representing correspondence with the above-discussed transformation functions is sufficient for most downstream tasks, but their necessity in representing correspondence (therefore efficiency) is application-dependent. This motivated this study to propose a different type of correspondence representation, using a set of corresponding ROIs. This is considered flexible to allow correspondence to be represented with only a few paired ROI masks, yet without losing generality (e.g. more and smaller ROI pairs) if denser local correspondence is required.     


For designing a learning-based registration algorithm to estimate corresponding ROI pairs, it is intuitive to use predefined ROI types and their segmented labels for training, akin to a weakly-supervised algorithm~\cite{hu2018weakly}, with or without intensity-based unsupervised loss~\cite{balakrishnan2019voxelmorph}. However, requiring ROI-labelled training data limits the general applicability of the algorithm. 

In this work, we propose a practical and general-purpose registration algorithm to overcome the limitation due to training data. Specifically, the registration process is streamlined using Segment Anything Model (SAM)~\cite{kirillov2023segment} to automatically generate prototypes from the multiple segmented ROIs on both images, and match them in pairs based on similarity. The presented experiments show that neither the definition nor the quantity of these SAM-segmented ROIs needs to be prescribed (or kept the same), for estimating sufficiently useful correspondences, in three different clinical image registration tasks.

In summary, our main contributions are: 
1)We pioneer a new correspondence representation for image registration using intuitive multiple paired-ROIs.
2) We introduce the general-purpose SAMReg by segmenting corresponding ROIs from both images, without any labels or fine-tuning. To our knowledge, it is the first application trial of SAM in image registration.
3) Extensive experiments show that our method outperforms the common-used registration tool and unsupervised registration, and is competitive with weakly-supervised registration that requires additional labeled training images, across three clinical applications.

\begin{figure}[!t]
    \centering
    \includegraphics[width=\textwidth]{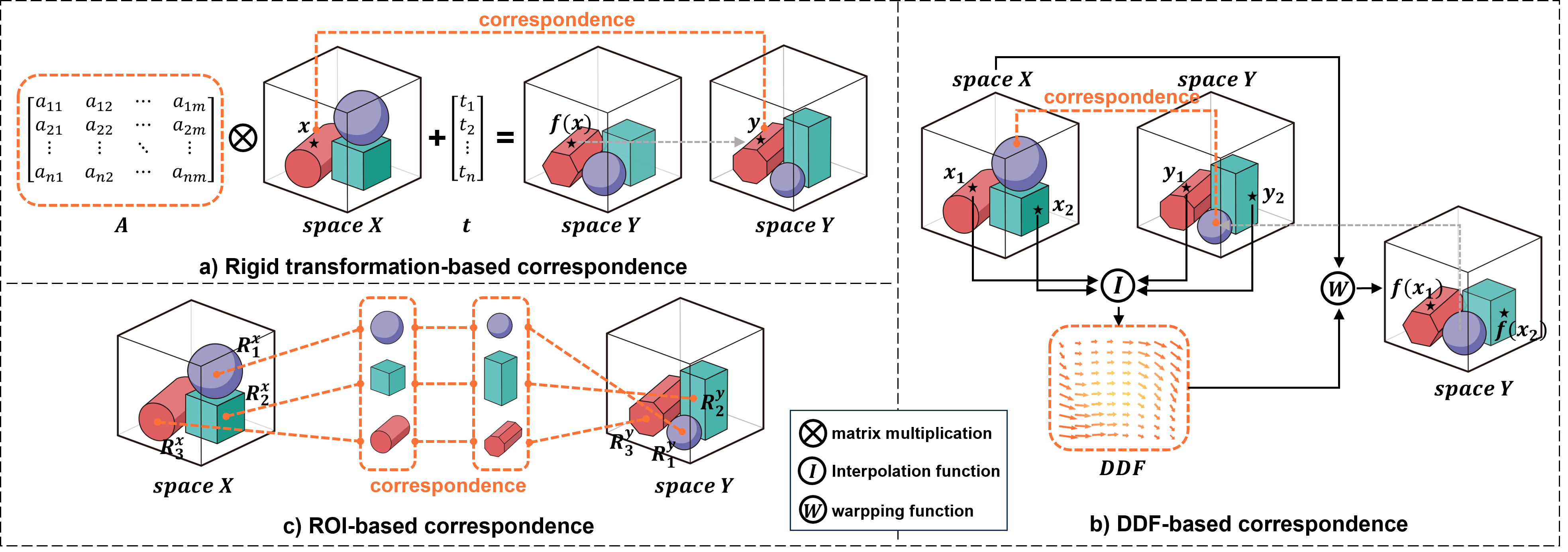}
    \caption{The correspondence representation diagram.}
    \label{fig:motivate}
\end{figure}

\section{ROI-based Correspondence Representation}
\label{sec:roi-correspondence}

This section considers spatial correspondence to be a mapping $f: \textbf{X} \rightarrow \textbf{Y}$ from spatial locations $\textbf{X}$ in a moving image coordinate system to locations $\textbf{Y}$ in the fixed image coordinate system. 
For example in 3D,  $\textbf{X}\in\mathbb{R}^{3}$ and $\textbf{Y}\in\mathbb{R}^{3}$ are random vectors containing Euclidean coordinates, in x-, y- and z-directions. 

Existing methods for representing spatial correspondence, $f$, fall into two main types: transformation functions of spatial locations or paired samples. The first type includes parametric functions like rigid, affine, and spline-based transformations (e.g. thin-plate-splines~\cite{rohr2001landmark})  that map input vectors $\textbf{x}$ to coordinate vectors $\textbf{y}$ through $\textbf{y}=f(\textbf{x}): \textbf{X} \rightarrow \textbf{Y}$. An example (shown in Fig.~\ref{fig:motivate}a) is the rigid transformation $f^{rigid}(\textbf{x})=A\textbf{x}+\textbf{t}$, where $A$ is a rotation matrix and $\textbf{t}$ is a translation vector. The second type represents correspondence through $N$ pairs of corresponding locations ${(\textbf{x}_n,\textbf{y}_n)}$, where $n=1,...,N$. For instance, DDF (shown in Fig.~\ref{fig:motivate}b) is represented by $\{(\textbf{x}_n,\textbf{x}_n+\textbf{d}_n)\}$ where $\textbf{d}_n$ is a set of vectors defined on all voxel locations $\textbf{x}_n$. Interpolation is applied to estimate $\textbf{y}$ for $\textbf{x}$ at locations beyond the defined voxels $\textbf{x}_n$. In this context, we formulate a more intuitive ROI-based correspondence representation as an alternative:

\begin{theorem}
Denoting $\textbf{x}$ and $\textbf{y}$ as spatial locations at respective moving- and fixed image spaces (i.e. coordinate systems), it is sufficient for $K$ pairs of regions-of-interest (ROIs), denoted as $\{(R^{x}_k, R^{y}_k)\}_{k=1}^{K}$, to indicate any spatial correspondence between the two image spaces, if $K$ is sufficiently large, where $R^{x}_k = \{\textbf{x}_l\}_{l=1}^{L^x_k}$ and $R^{y}_k = \{\textbf{y}_l\}_{l=1}^{L^y_k}$ are two sets of $L^x_k$ and $L^y_k$ spatial locations that represent the same (corresponding) ROIs, in the moving- and fixed image spaces, respectively.
\end{theorem}

This is intuitive in an extreme case, in which a large number ($K=N$) of single-voxel (i.e. $L^x_k=L^y_k=1$) ROIs are used, the ROI pairs can be sampled at $N$ voxels to an equivalent DDF representation $\{(R^{x}_k = \textbf{x}_k, R^{y}_k=\textbf{y}_k)\}_{k=1}^{N}$. 

Therefore, an informal proof for Theorem 1 can be understood by iterating the following process until sufficiency is reached:
When a point-to-point correspondence $\textbf{y}^*=f(\textbf{x}^*)$ that has NOT been represented by current ROI pairs $\{(R^{x}_k, R^{y}_k)\}_{k=1}^{K}$, one can always sample an additional ROI pair $\{(R^{x}_{K+1}=\textbf{x}^*, R^{y}_{K+1}=\textbf{y}^*)\}$, such that the resulting ($K+1$) ROI pairs $\{(R^{x}_k, R^{y}_k)\}_{k=1}^{K+1} = \{(R^{x}_k, R^{y}_k)\}_{k=1}^{K} \cup \{(R^{x}_{K+1}, R^{y}_{K+1})\}$ suffice.

Further, properties of the proposed ROI-based correspondence are summarized: 1) individual ROIs can be practically depicted as binary pair masks; 2) paired ROIs can represent both local and global correspondence; 3) Overlapping ROIs facilitate one-to-many correspondence; 4) if dense-level correspondence is required, the correspondence at any spatial location can be inferred. We discuss specific algorithms in Sec.~\ref{sec:align}; 5) if dense-level correspondence is not required, omitting smoothness regularization (e.g., rigidity or bending energy) during estimation enhances registration efficiency with ROI-based representation.

\section{One Registration versus Two Segmentations}
\label{One registration}

Given a single-image $I^{x}$, a multi-class segmentation is to estimate the joint probability of a random vector $\textbf{C}^x=[C^x_1,...,C^x_{K^x}]^\top\in\mathbb{R}^{K^x}$, representing $K^x$-class probabilities over $N$ voxel locations, $\prod_{n=1}^{N} p_n(\textbf{C}^x | I^{x})$. The segmented ROI set $\{\hat{R}^{x}_k\}_{k=1}^{K^x}$ can be computed by binarising the estimated class probability vectors $\{\hat{\textbf{c}}_n^x\}^{N}_{n=1}$, where $\hat{\textbf{c}}_n^x \sim \textbf{C}^x$. 
For a second image $I^{y}$, ROIs ${\hat{R}^{y}_k}$ are segmented by estimating $K^y$-class probabilities ${\hat{\textbf{c}}_n^y}$, where $\hat{\textbf{c}}_n^y$ is the class probabilities for $K^y$ classes.
In this section, we consider the following two cases for registering $I^{x}$ and $I^{y}$, by estimating ROI-based correspondence described in Sec.~\ref{sec:roi-correspondence}.

\textbf{Segmenting the same ROIs from two images.}
In this case, the segmentation algorithm (e.g. one or two neural networks) is capable of segmenting the same classes of ROIs, i.e. $\textbf{C}=\textbf{C}^x=\textbf{C}^y$ and $K=K^x=K^y$, from both images. Therefore, the obtained ROIs pairs can directly represent the correspondence, described in Sec.~\ref{sec:roi-correspondence}, $\{(R^{x}_k=\hat{R}^{x}_k, R^{y}_k=\hat{R}^{y}_k)\}_{k=1}^{K}$, that is to estimate the monotonic joint probability given two images $p_n(\textbf{C} | I^{x},I^{y}) \propto p_n(\textbf{C} | I^{x}) \cdot p_n(\textbf{C} | I^{y})$, subject to normalization constants. This approach usually requires predefined ROI types and segmentation networks trained using these ROI-labelled data. Interestingly, these segmented training data are exactly the same data required for training weakly-supervised registration algorithms~\cite{hu2018weakly,hu2019conditional}.

\textbf{Matching the segmented ROI candidates.}
Assume an alternative "unsupervised" segmentation can be used to obtain two sets of candidate ROIs, $\{\hat{R}^{x}_k\}_{k=1}^{K^x}$ and $\{\hat{R}^{y}_k\}_{k=1}^{K^y}$, from the fixed and moving- images, respectively. Search for a subset of common ROI classes, $\textbf{C}=[C_1,...,C_K]^\top$, where $\{C_k\}_{k=1}^K = \{C^x_k\}_{k=1}^{K^x} \cap \{C^y_k\}_{k=1}^{K^y}$. Thus, the corresponding ROI pairs $\{(R^{x}_k, R^{y}_k)\}_{k=1}^{K}$ derived from the estimated probability vectors $\hat{\textbf{c}}_n \sim \textbf{C}$, that is the joint probability between the two conditionally independent "candidate probabilities", $p_n(\textbf{C} | I^{x},I^{y}) = p_n(\textbf{C}^x,\textbf{C}^y | I^{x},I^{y}) = p_n(\textbf{C}^x | I^{x},I^{y}) \cdot p_n(\textbf{C}^y | I^{x},I^{y}) \propto p_n(\textbf{C}^x | I^{x}) \cdot p_n(\textbf{C}^y | I^{y})$. In the following Sec.~\ref{sec:seg_match}, we propose a specific algorithm using this strategy.



\section{Segment Everything and Match: A Training-Free Registration Algorithm}
\label{sec:seg_match}

\subsection{Preliminary: SAM Architecture}
SAM~\cite{kirillov2023segment} 
consists of an image encoder, a prompt encoder and a mask decoder, is adept at image segmentation without task-specific training, and has recently been adapted for medical imaging. The adaptation includes training on large medical datasets~\cite{ma2024segment,cheng2023sammed2d}, thus establishing a foundation model, and fine-tuning for targeted medical datasets using, for example, additional adaptation layers~\cite{wu2023medical,gong20233dsam,wang2023sammed3d}. Despite the advancements, as highlighted by these authors, SAM may yields limited sensitivity to medical images compared to natural images~\cite{huang2024segment,mazurowski2023segment}, resulting in mixed results in direct applications for specific anatomy and pathology segmentation without further training. 


It is interesting to argue that requiring known anatomical significance in segmentation may not limit medical image registration. Although correspondence between medical images may be application-dependent to some extent~\cite{crum2003zen}, the main objective is to find \textit{similar} regions across images, a task that suits SAM's strong feature and pixel classification capabilities. For example, SAM can effectively segment ROIs with clear edges such as the iliac internal, which are not typical targets but useful for establishing local correspondence. By deploying SAM to segment extensive foreground areas, we aim to discover corresponding ROIs that capture diverse anatomical structures and areas that may lack precise anatomical definitions (\textit{e.g.}, common cavities between structures).


\subsection{The SAMReg Algorithm}
\label{sec:align}
\begin{figure}[!t]
    \centering
    \includegraphics[width=\textwidth]{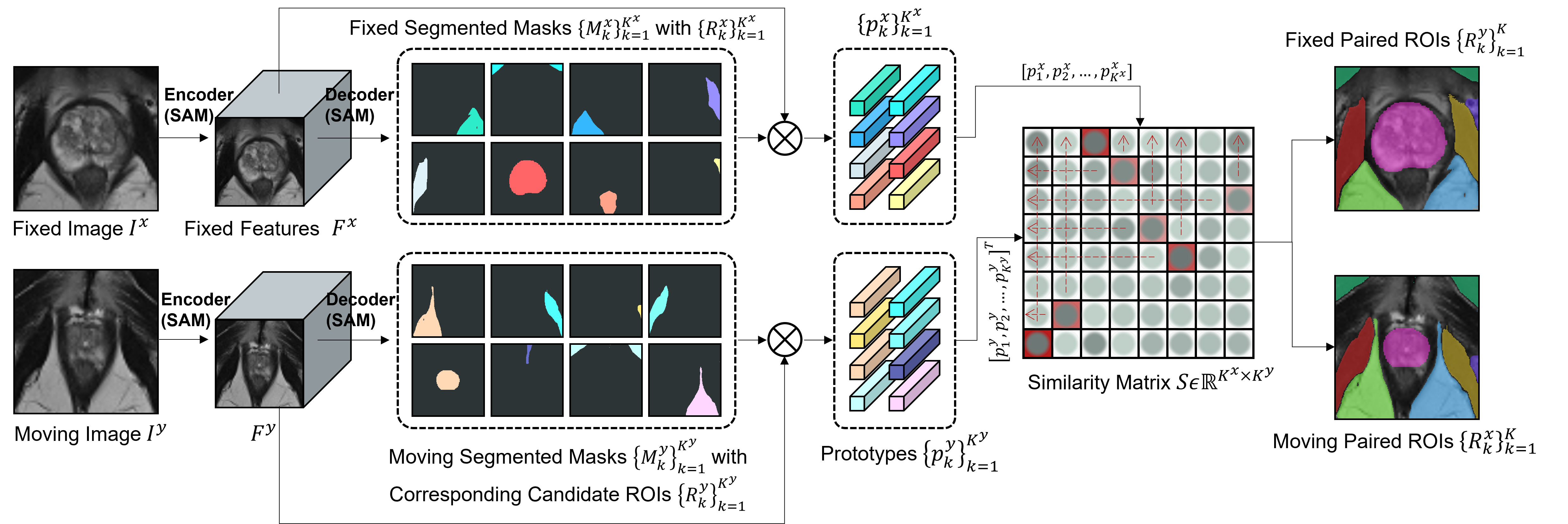}
    \caption{The pipeline of the proposed SAM-Reg method.}
    \label{fig:pipeline}
\end{figure}

Since the multiple ROIs generated by SAM segmentation lack explicit one-to-one correspondence, our strategy involves embedding these ROIs before matching them in feature space, with the proposed pipeline illustrated in Fig.~\ref{fig:pipeline}.

\textbf{ROI Embedding:} Leveraging the SAM, we independently encode the input images $I^{x}$ and $I^{y}$ into features, $F^{x}$ and $F^{y}$ respectively. By setting the SAM decoder to the \textit{everything} mode with a simple outlier-removing filter (further details in Sec.~\ref{sec:exp_results}), we exploit its generality to produce a comprehensive array of segmentation masks for each image, obtaining mask sets $\{M_k^{x}\}_{k=1}^{K^x}$ and $\{M_k^{y}\}_{k=1}^{K^y}$, representing ROIs $\{R_k^{x}\}_{k=1}^{K^x}$ and $\{R_k^{y}\}_{k=1}^{K^y}$, respectively. 

For each binary mask, we derive moving prototypes $\{p_k^{x}\}_{k=1}^{K^x}$ by element-wise multiplying the mask with their corresponding feature map, element being indexed by $n$:  
$p^x_k = \frac{\sum_{n}M^{x,res}_k(n)\cdot F^x(n)}{\sum_{n}M^{x,res}_k(n)}$, 
where $res$ indicates resizing $M^{x}_k$ to the size of $F^x$ as $M^{x,res}_k$. This also applies on the fixed prototypes $\{p_k^{y}\}_{k=1}^{K^y}$.

\textbf{ROI Matching:} Assume a similarity matrix $S \in \mathbb{R}^{K^x \times K^y}$ to measure the cosine similarity between the moving- and fixed prototypes:
$S(i,j)=\lVert\frac{p_i^{x}\cdot p_j^{y}}{\lVert p_i^{x} \rVert \cdot \lVert p_j^{y} \rVert}\rVert$, where $\lVert\cdot\rVert$ normalises each element $S(i,j)\in[0,1]$, $i\in[1,K^x]$ and $j\in[1,K^y]$.

A set of index pairs $P$ then can be identified: $P = \mathop{\arg\max}\limits_{i,j}(S(i,j))$, \textit{s.t.}~ $S(i,j)>\epsilon ~\forall~{(i,j)}$, corresponding to pairings with maximum similarities. A further constraint is added to ensure each $i, j$ is chosen at most once, if overlapping ROIs segmented from SAM are filtered. $\epsilon$ is the similarity threshold and the set length $K = |P| \leq min(K^x,K^y)$.
Using the selected index pairs $P$, we construct two new sets of masks, $\{M_k^{x,cor}\}_{k=1}^{K}$ and $\{M_k^{y,cor}\}_{k=1}^{K}$:
\begin{equation}
  (M_k^{x}, M_k^{y})^{cor} = \{(M_i^{x}, M_j^{y}) | (i, j) \in P\}, k=1,2,...,K.  
\end{equation}
This leads to the final paired ROIs indicating the correspondence between $I^{x}$ and $I^{y}$. This matching mechanism can be applied on either 2D images or 3D images, in the latter of which, SAM-segmented 2D slices can be stacked to provide the candidate masks, $\{M_k^{x}\}_{k=1}^{K^x}$ and $\{M_k^{y}\}_{k=1}^{K^y}$ in 3D, before computing prototypes.

\textbf{Optional ROI-to-Dense Correspondence:} Further, as discussed in Sec.~\ref{One registration}, we explore the capability of converting estimated ROI-based correspondence to its dense counterpart DDF, useful for applications such as full-image alignment. In this section, we describe a general method that iteratively refines the DDF, to optimize an objective function $\mathcal{L}$ that combines a region-specific alignment measure $\mathcal{L}_{roi}$ with a regularization term $\mathcal{L}_{ddf}$ to ensure smooth interpolation:

\begin{equation}
\mathcal{L}_{\Theta}(\{(M_k^{x}, M_k^{y})\}) = \sum_{k=1}^K \mathcal{L}_{roi}(M_k^{x}, \mathcal{T}(M_k^{y},\Theta)) + \lambda \mathcal{L}_{ddf}(\Theta) 
\end{equation}

where $\Theta$ represents parameters of the transformation function $\mathcal{T}$ and $\lambda$ is a regularization parameter that balances the alignment accuracy with the deformation smoothness. This may itself be considered a multi-ROI alignment algorithm but with known ROI correspondence. In this work, we use an equally-weighted MSE and Dice as $\mathcal{L}_{roi}$ and a L$^2$-norm of DDF gradient as $\mathcal{L}_{ddf}$. 


\section{Experiments and Results}
\label{sec:exp_results}
\begin{table}
\scriptsize
\centering
\caption{Comparison with the-state-of-the-art medical image registration methods, where the $*$  indicates methods specifically trained on the dataset.}
\label{tab:SOTA}
\setlength{\tabcolsep}{1.5pt}
 \renewcommand{\arraystretch}{1.2} 
\begin{tabular}{c|cc|cc|cc} 
\hline
\multirow{2}{*}{\textbf{Method}} & \multicolumn{2}{c|}{\textbf{MR-Prostate~\cite{ahmed2017diagnostic}}} & \multicolumn{2}{c|}{\textbf{MR-Cardiac~\cite{bernard2018deep}}} & \multicolumn{2}{c}{\textbf{CT-Abdomen~\cite{ji2022amos}}} \\ 
\cline{2-7}
 & \textbf{Dice} & \textbf{TRE} & \textbf{Dice} & \textbf{TRE} & \textbf{Dice} & \textbf{TRE} \\ 
\hline
NiftyReg~\cite{modat2014global} & 7.68±3.98 & 4.67±3.48 & 9.93±2.21 & 3.29±2.89 & 6.81±3.02 & 5.86±3.64 \\
VoxelMorph*~\cite{balakrishnan2019voxelmorph} & 56.84±3.41 & 3.68±1.92 & 60.10±3.95 & 3.03±2.41 & 58.34±3.72 & 4.13±2.23 \\
LabelReg*~\cite{hu2018weakly} & 77.32±3.56 & 2.72±1.23 & 78.97±2.42 & 1.73±1.34 & 74.53±3.43 & 2.58±1.18 \\
SAMReg(Ours) & 75.67±3.81 & 2.09±1.35 & 80.28±3.67 & 1.43±1.13 & 71.74±3.72 & 2.48±1.01 \\
\hline
\end{tabular}
\end{table}
\vspace{20pt}
\begin{table}
\centering
\caption{Ablation study of segmentation performance: SAM versus leading dataset-specific segmentation models. \textit{Lab-ROI} represents the dataset's original labeled ROIs, while \textit{Pse-ROI} refers to pseudo random ROIs identified by SAM.}
\label{tab:seg_model}
\scriptsize
\setlength{\tabcolsep}{1.5pt}
 \renewcommand{\arraystretch}{1.2} 
\begin{tabular}{c|c|c|c|c|c} 
\hline
\multirow{2}{*}{\textbf{Dataset}} & \multirow{2}{*}{\textbf{Seg.Model}} & \multicolumn{2}{c|}{\textbf{Lab-ROI}} & \multicolumn{2}{c}{\textbf{Pse-ROI}} \\ 
\cline{3-6}
 &  & \textbf{Dice} & \textbf{TRE} & \textbf{Dice} & \textbf{TRE} \\ 
\hline
\multirow{2}{*}{Prostate~\cite{ahmed2017diagnostic}} & SAM~\cite{kirillov2023segment} & 75.67±3.81 & 2.72±1.23 & 97.73±3.15 & 0.82±0.23 \\
 & Vnet~\cite{milletari2016v} & 80.35±3.14 & 2.13±1.24 & 43.56±3.54 & 5.32±2.51 \\ 
\hline
\multirow{2}{*}{Cardiac~\cite{bernard2018deep}} & SAM~\cite{kirillov2023segment} & 80.28±3.67 & 1.73±1.34 & 98.87±3.61 & 1.62±0.67 \\
 & FCT~\cite{tragakis2023fully} & 83.54±3.46 & 1.31±1.25 & 31.35±3.74 & 7.24±2.52 \\ 
\hline
\multirow{2}{*}{Abdomen~\cite{ji2022amos}} & SAM~\cite{kirillov2023segment} & 71.74±3.72 & 2.58±1.18 & 98.24±3.17 & 1.45±0.75 \\
 & MedNeXt~\cite{roy2023mednext} & 78.26±3.34 & 1.84±1.54 & 45.73±3.73 & 5.37±2.12 \\
\hline
\end{tabular}
\end{table}

\begin{figure}[!t]
    \centering
    \includegraphics[width=\textwidth]{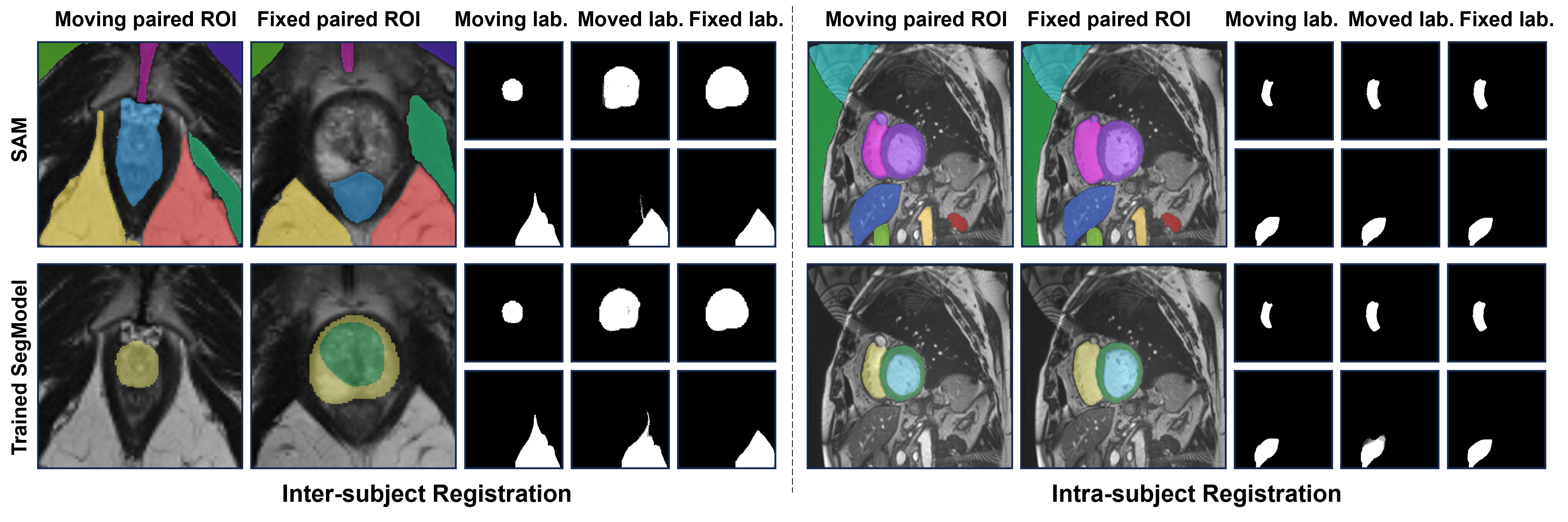}
    \caption{The qualitative comparisons of SAM and fully-supervised segmentation results, including ROI-label(top) and pseudo-label(bottom)wrapping outcomes.}
    \label{fig:vis}
\end{figure}

\noindent\textbf{Datasets and Implementation Details:}
Our method is evaluated on MR-Prostate~\cite{ahmed2017diagnostic}, MR-Cardiac~\cite{bernard2018deep}, and CT-Abdomen~\cite{ji2022amos} datasets, highlighting diverse anatomical areas and modalities. We apply inter-subject registration for Prostate and Abdomen datasets to align images to a standard reference, and intra-subject registration for Cardiac images, aligning time-variant image pairs.
Performance is measured using the Dice score for overlap and target registration error (TRE) for landmark alignment, where TRE is defined as the root-mean-square (RMS) distance. For non-rigid registration~\cite{eppenhof2018deformable,sotiras2013deformable}, we assess accuracy by the mean transformation accuracy of critical anatomical structures~\cite{hu2018weakly}, with TRE calculated from the centroids of the relevant ROIs.
In configuring the SAM parameters, beyond the standard settings, we adjusted the pred-iou-thresh and stability-score-thresh to 0.90. Our ROI filtering strategy is based on area and overlapping ratio, setting a minimum of 200 and a maximum of 7000 for the ROI sizes and a maximum ratio of 0.8. The similarity threshold $\epsilon$ is set to 0.8. Our network is implemented with pytorch and MONAI~\cite{cardoso2022monai}. Code demo has been released at: \url{https://github.com/sqhuang0103/SAMReg.git}.

\begin{table}
\begin{minipage}{0.5\textwidth}
\caption{Ablation of paired ROI quantity on MR-prostate dataset.}
\label{tab:ROI_number}
\centering
\scriptsize
\setlength{\tabcolsep}{1.5pt}
 \renewcommand{\arraystretch}{1.2} 
\begin{tabular}{c|cc} 
\hline
\textbf{PairNum} & \textbf{Dice} & \textbf{TRE} \\ 
\hline
1 & 69.59±3.51 & 3.65±1.53 \\
2 & 74.19±3.41 & 2.63±1.76 \\
3 & 76.00±3.23 & 2.02±1.35 \\
4 & 75.83±3.43 & 2.42±1.65 \\
All & 74.67±3.31 & 2.72±1.53 \\
\hline
\end{tabular}
\end{minipage}\quad
\begin{minipage}{0.5\textwidth}
\caption{Ablation of minimum similarity threshold $\epsilon$  on MR-prostate dataset.}
\label{tab:threshold}
\centering
\scriptsize
\setlength{\tabcolsep}{1.5pt}
 \renewcommand{\arraystretch}{1.2} 
\begin{tabular}{c|cc} 
\hline
\textbf{Threshold $\epsilon$} & \textbf{Dice} & \textbf{TRE} \\ 
\hline
0.2 & 72.76±3.63 & 3.65±1.38 \\
0.4 & 74.43±3.40 & 2.46±1.56 \\
0.6 & 75.78±3.24 & 2.20±1.43 \\
0.8 & 70.43±3.18 & 3.13±1.53 \\
\hline
\end{tabular}
\end{minipage}
\end{table}

\noindent\textbf{Comparing SAMReg with SOTA Methods:}
In our evaluation, we fairly compare SAMReg with NiftyReg~\cite{modat2014global}, a non-learning tool, and also contrast it, less equitably, with tailored methods like VoxelMorph~\cite{balakrishnan2019voxelmorph} and LabelReg~\cite{hu2018weakly}. VoxelMorph and LabelReg are fine-tuned on the research datasets, employing unsupervised and weakly-supervised learning respectively, with LabelReg also depending on anatomical annotations. Despite this, SAMReg demonstrates outstanding performance without dataset-specific training, particularly in the intra-subject cardiac dataset due to consistent patterns, as illustrated in Table~\ref{tab:SOTA}. For the inter-subject datasets of MR-Prostate and MR-Cardiac, SAMReg slightly lags behind in Dice scores but showcases competitive TRE performance, highlighting its interesting local alignment ability over shape conformity, perhaps due to the involvement of multiple ROIs. Furthermore, the ability to achieve such results without dataset-specific training underscores the robust generalization capacity of SAMReg, suggesting its wide applicability in medical tasks, and reducing the time and resources needed for clinical deployment.

\noindent\textbf{Ablative Study:}

\textbf{\textit{ROI Segmentation Comparison:}} 
Table~\ref{tab:seg_model} and Fig.~\ref{fig:vis} present ROI segmentation and the registration alignment performance comparisons between SAM and top segmentation models. Despite an unfair advantage for benchmark models designed and trained for groundtruth annotations (Label-ROI), SAMReg demonstrates robust performance and effectiveness, without ground-truth reliance, even outperforming these models with SAM-generated pseudo ROIs (Pse-ROI). 

\textbf{\textit{Optimal Number $K$ of ROI Pairs:}}
Table~\ref{tab:ROI_number} shows that using up to three ROI pairs optimizes registration, improving Dice and TRE scores. This is an interestingly small number, suggesting future work for even furthering registration performance, by more effectively utilizing smaller paired ROIs.



\textbf{\textit{Threshold $\epsilon$ for ROI Correspondence:}}
As illustrated in Table~\ref{tab:threshold}, a 0.6 similarity threshold enhances Dice scores and reduces TRE, while a higher threshold (0.8) reduces ROI selection and performance, underscoring the importance of a balanced threshold for optimal registration results.



\section{Conclusion}
In this study, We introduce a novel ROI-based correspondence representation. With this representation, the image registration is reformalized as two multi-class segmentation tasks, with a proposed, general-purpose and practical implementation, SAMReg. The comprehensive experiments show competitive registration performance that promises a new research direction in image registration.

\subsubsection{\ackname}
This work was supported by the International Alliance for Cancer Early Detection, a partnership between Cancer Research UK [C28070/A30912; C73666/A31378], Canary Center at Stanford University, the University of Cambridge, OHSU Knight Cancer Institute, University College London and the University of Manchester.


\bibliographystyle{splncs04_unsort}
\bibliography{refs}

\end{document}